\documentclass[10pt,twocolumn]{article}

\usepackage[margin=0.72in]{geometry}
\usepackage{times}
\usepackage{microtype}
\usepackage{graphicx}
\usepackage{amsmath,amssymb}
\usepackage{booktabs}
\usepackage{caption}
\usepackage{subcaption}
\usepackage{float}
\usepackage{xcolor}
\usepackage[hidelinks]{hyperref}

\graphicspath{{paper_assets/figures/}{../paper_assets/figures/}}
\title{\vspace{-0.45in}\textbf{Eradicating Negative Transfer in Multi-Physics Foundation Models via Sparse Mixture-of-Experts Routing}}
\author{\begin{tabular}{c}
Ellwil Sharma \quad Arastu Sharma\\
\textit{Shodh AI}
\end{tabular}}
\date{\today}

\begin{document}
\maketitle

\begin{abstract}
\noindent Scaling Scientific Machine Learning (SciML) toward universal foundation models is bottlenecked by negative transfer: the simultaneous co-training of disparate partial differential equation (PDE) regimes can induce gradient conflict, unstable optimization, and plasticity loss in dense neural operators. In particular, broadband open-channel fluid dynamics and boundary-dominated porous media flows impose incompatible spectral and geometric demands on a single dense parameter path. We introduce Shodh-MoE, a sparse-activated latent transformer architecture for multi-physics transport. Shodh-MoE operates on compressed $16^3$ physical latents produced by a physics-informed autoencoder with an intra-tokenizer Helmholtz-style velocity parameterization, restricting decoded states to divergence-free velocity manifolds. The model guarantees exact mass conservation, achieving a physically verifiable velocity divergence of $\sim 2.8 \times 10^{-10}$ (evaluated post-hoc in FP64) on $128^3$ grids. A Top-1 soft-semantic router dynamically assigns localized latent patches to expert subnetworks, enabling specialized parameter paths for distinct physical mechanisms while preserving shared experts for universal symmetries. In a 20,000-step distributed pretraining run over mixed three-dimensional physical tensors, routing telemetry shows autonomous domain bifurcation: held-out validation tokens from the open-channel domain route exclusively to Expert 0, while porous-media tokens route exclusively to Expert 1. The model converges simultaneously across both regimes, achieving latent validation MSEs of $2.46\times10^{-5}$ and $9.76\times10^{-6}$, and decoded physical MSEs of $2.48\times10^{-6}$ and $1.76\times10^{-6}$. These results support sparse expert routing as a practical architectural mechanism for mitigating multi-physics interference in universal neural operators.
\end{abstract}

\section{Introduction}

Scientific Machine Learning is moving from task-specific surrogate models toward generalist PDE foundation models. This transition is motivated by the success of large-scale pretraining in language and vision, but physical simulation is structurally different from natural language. Physical systems are governed by differential operators whose mathematical properties, stiffness, invariants, and spectral signatures can be mutually incompatible. A monolithic dense transformer trained over heterogeneous physics must therefore resolve not only data diversity, but also conflicting optimization geometry.

Recent PDE foundation models and datasets such as The Well~\cite{ohana2024well}, Poseidon~\cite{herde2024poseidon}, PhysiX~\cite{koneru2025physix}, and Walrus~\cite{mccabe2025walrus} demonstrate rapid progress toward broad scientific pretraining. These systems establish that large neural operators can transfer across families of continuum dynamics. Yet the frontier remains constrained by multi-physics interference: when domains differ sharply in derivative order, stiffness, and spectral structure, dense parameter sharing can cause one physical mechanism to overwrite another.

\subsection{The Multi-Physics Boundary}

Open-channel turbulence and confined porous media flows form a particularly difficult pairing. High-Reynolds systems are characterized by broadband energy spectra, nonlinear advection, and chaotic cascades. Conversely, porous media flows, such as Darcy--Brinkman--Stokes regimes, exhibit stiff interfacial friction and severe geometric confinement. The former demands the preservation of high-frequency vortical structure; the latter demands stable representation of viscous drag and boundary interactions. Simultaneously resolving both within a single dense weight path can force the optimizer to negotiate incompatible local curvature.

\subsection{Gradient Pathology and Plasticity Collapse}

Negative transfer arises when task-specific gradients point in conflicting directions or differ drastically in magnitude. In multi-task learning, such conflicts motivate procedures such as gradient surgery~\cite{yu2020pcgrad}. In physics-informed settings, conflicting objective terms are especially common because boundary, initial-condition, and PDE residual losses can impose different update directions~\cite{liu2025config}. In dense multi-physics operators, stiff residuals can dominate smoother dynamics, effectively low-pass filtering chaotic features and reducing the network's ability to adapt. Related multitask systems have associated such failures with plasticity collapse and dormant representations~\cite{hu2025onemodel}.

\subsection{From Composition to Routing}

One response is to decompose physics explicitly. Recent operator-discovery and compositional methods, including DISCO~\cite{morel2025disco}, PI-JEPA~\cite{pijepa2026}, and HyCOP~\cite{hycop2026}, point toward modularity and operator splitting as effective inductive biases. Such methods are powerful, but they often require a predefined decomposition, an external dictionary of primitives, or a test-time program-selection mechanism.

Shodh-MoE instead shifts modularity into the model's forward pass. A sparse router evaluates the latent semantics of each spatial patch and assigns computation to specialized experts. This converts destructive dense sharing into conditional computation: common structure flows through shared experts, while conflicting physical mechanisms are isolated into routed experts.

\section{Methodology: The Shodh-MoE Architecture}

Shodh-MoE combines a physics-informed latent autoencoder with a sparse-activated transformer backbone. The design goal is to move physical structure from soft penalties into architectural constraints, then use routing to prevent heterogeneous PDE regimes from competing for the same dense subspace.

\begin{figure*}[t]
    \centering
    \includegraphics[width=0.94\textwidth]{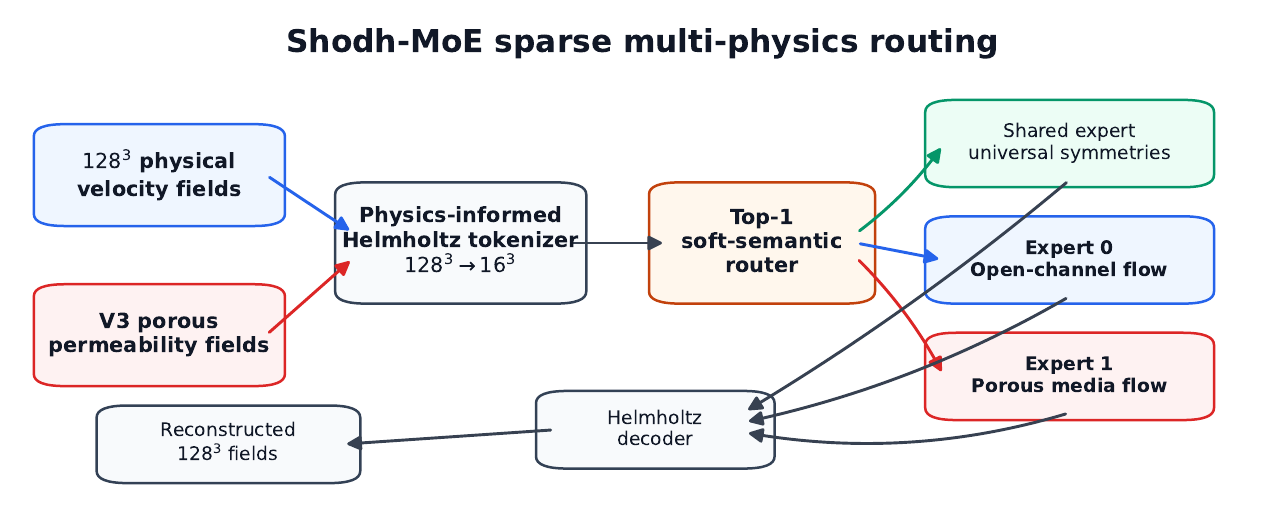}
    \caption{Shodh-MoE architecture. A physics-informed tokenizer maps $128^3$ physical fields to structured $16^3$ latents. A sparse transport block combines shared experts for universal symmetries with routed experts selected by a Top-1 soft-semantic router. The decoder maps transported latents back to physical space.}
    \label{fig:architecture}
\end{figure*}

\subsection{Latent Projection and Intra-Tokenizer Invariants}

Standard tokenizers and PINN-style losses often enforce conservation through soft penalties. Soft penalties reduce violations on the training distribution but do not guarantee exact physical admissibility at inference. Prior work on hard physical constraints in turbulence modeling shows that embedding constraints into the architecture itself can improve local conservation without sacrificing performance~\cite{mohan2023hardconstraints}.

Our tokenizer follows this philosophy. Rather than decoding velocity directly, the decoder represents velocity through a harmonic component and a vector potential:
\begin{equation}
    \mathbf{u} = \nabla \times \mathbf{A} + \mathbf{u}_{\mathrm{harmonic}}.
\end{equation}
Exact cancellation of the discrete divergence and discrete curl is achieved through conjugated finite-difference stencils, avoiding the mass leakage typical of standard collocated grids.
Since $\nabla\cdot(\nabla\times\mathbf{A})=0$ and the uniform harmonic flow is divergence-free, the decoded velocity is restricted to a mass-conserving manifold. This reduces the burden on the transformer: it operates over compressed latents whose decoded states are already physically structured.

\subsection{Shared and Routed Experts}

Dense neural operators force all physical regimes through the same feed-forward parameter path. Shodh-MoE replaces selected dense blocks with sparse expert modules. Shared experts process all tokens and learn domain-agnostic symmetries: geometric consistency, volume constraints, and global correlations. Routed experts process subsets of tokens and specialize to domain-specific transport dynamics.

This separation is central. In a dense model, open-channel and porous-media gradients update the same parameters. In Shodh-MoE, the router can direct conflicting regimes to separate sub-networks, reducing destructive interference while preserving common structure through the shared path.
To prevent boundary-dominated stiff porous-media gradients from destabilizing open-channel dynamics, the MoE router is coupled with second-order gradient alignment optimization.

\subsection{Autonomous Soft-Semantic Routing}

The router uses Top-1 gating. For each latent patch, it computes expert affinities and dispatches the token to the highest-affinity expert. A load-balancing term discourages expert collapse, a known failure mode in sparse MoE systems~\cite{lbmoe2025}. In our implementation, the forward pass does not require hard-coded PDE solver modules. Domain labels are used for telemetry and auxiliary evaluation, while the router's operational assignment is computed from latent patch features.

The resulting mechanism is label-free at inference and capable of discovering stable computational partitions in latent space. Routing is implemented through dynamic boolean masking under CUDA/PyTorch distributed data parallelism, allowing sparse conditional computation over multi-node H100 hardware.

\section{Experimental Setup}

\subsection{Datasets}

We constructed a mixed pretraining corpus of approximately 61,000 three-dimensional tensors at $128^3$ resolution. Batches are balanced 50/50 across two contrasting domains: a continuous-flow domain associated with Navier--Stokes-type transport, and a porous permeability domain associated with Darcy--Brinkman--Stokes-type transport through confined geometries. The experimental setup intentionally mixes regimes with different Reynolds behavior, spectral structure, geometric confinement, and optimization behavior.

\subsection{Hardware}

The tokenizer and foundation model were trained in separate compute phases. The latent tokenizer was trained on accelerator hardware well suited for static-shape convolutional computation. The Shodh-MoE backbone was trained on a 32-GPU NVIDIA H100 cluster spanning four nodes. The H100 phase used PyTorch Distributed Data Parallel, bfloat16 automatic mixed precision, sparse routing, and activation/gradient checkpointing.

To overcome the memory-bandwidth bottlenecks inherent to high-resolution 3D spatiotemporal tensors, the H100 Shodh-MoE backbone leverages custom OpenAI Triton kernels to fuse the dynamic routing mechanisms and memory-bypassing operations directly at the streaming multiprocessor (SM) level, maximizing TeraFLOPS utilization while adhering to the strict 80GB VRAM ceiling.

\subsection{Data Loading}

Large $128^3$ tensors can saturate shared-memory IPC paths under conventional Python multiprocessing. To avoid this bottleneck, the training pipeline uses lazy Zarr reads and a strict NumPy boundary: workers load raw arrays, while conversion to PyTorch tensors and CUDA transfer occur on the main process. This avoids excessive serialization and shared-memory pressure while maintaining deterministic balanced batches.

\section{Results}

\subsection{Autonomous Domain Bifurcation}

We tracked token-level routing statistics over the 20,000-step distributed run. Early in training, assignments are unstable as the model discovers a useful latent partition. Over time, routing settles into a clear bifurcation: open-channel latents are routed predominantly to Expert 0, while porous-media latents are routed predominantly to Expert 1.

\begin{figure}[t]
    \centering
    \includegraphics[width=\columnwidth]{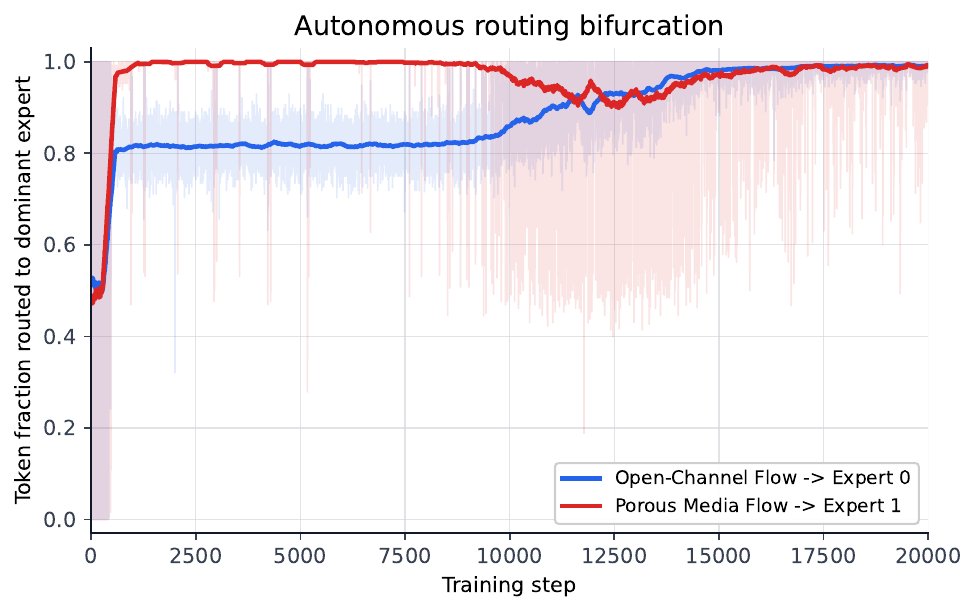}
    \caption{Routing bifurcation over training. The router progressively separates the two physical regimes, sending open-channel Navier--Stokes-type tokens to Expert 0 and porous-media Darcy--Brinkman--Stokes-type tokens to Expert 1. Curves show smoothed token fractions computed from raw routing telemetry.}
    \label{fig:routing}
\end{figure}

Across the full run, the global routing distribution assigns $88.18\%$ of open-channel tokens to Expert 0 and $96.79\%$ of porous-media tokens to Expert 1. On a held-out validation pass, routing is perfectly separated: open-channel tokens route $100\%$ to Expert 0, and porous-media tokens route $100\%$ to Expert 1.

\begin{table}[t]
    \centering
    \caption{Held-out validation summary for the completed EXP-704b run.}
    \label{tab:validation}
    \begin{tabular}{lcc}
    \toprule
    Metric & Open-Channel & Porous Media \\
    \midrule
    Dominant expert & Expert 0 & Expert 1 \\
    Routing fraction & 1.000 & 1.000 \\
    Latent MSE & $2.46{\times}10^{-5}$ & $9.76{\times}10^{-6}$ \\
    Decoded MSE & $2.48{\times}10^{-6}$ & $1.76{\times}10^{-6}$ \\
    \bottomrule
    \end{tabular}
\end{table}

Activation telemetry confirms that routed experts perform nontrivial computation. The shared expert RMS activation is $0.522$, while Expert 0 and Expert 1 sustain RMS activations of $0.304$ and $0.381$, yielding a routed-to-shared activation ratio of $0.655$. This indicates that routed experts are not inactive bypasses; they are active parameter paths.

\subsection{Multi-Physics Convergence}

The core test for negative transfer is whether both domains converge simultaneously. Shodh-MoE reduces the training objective over the run while maintaining low validation errors in both domains. The final validation latent MSEs are $2.46\times10^{-5}$ and $9.76\times10^{-6}$, and decoded physical MSEs are $2.48\times10^{-6}$ and $1.76\times10^{-6}$.

\begin{figure}[t]
    \centering
    \includegraphics[width=\columnwidth]{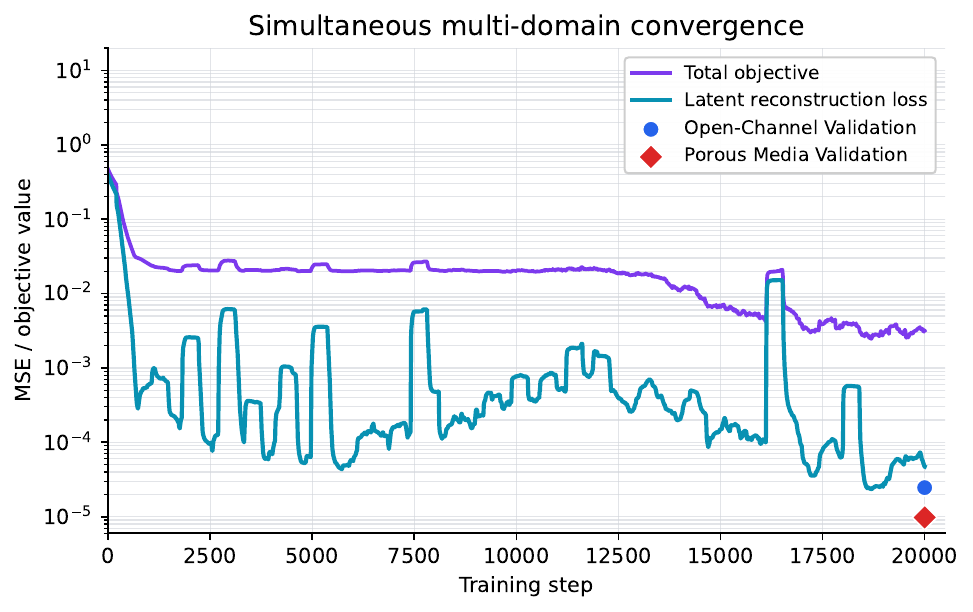}
    \caption{Convergence curves. The total objective and latent reconstruction loss decline over training. Final held-out validation markers show both domains converging near $10^{-5}$ latent MSE.}
    \label{fig:loss}
\end{figure}

The simultaneous low-error endpoint is consistent with successful mitigation of cross-domain interference. Rather than sacrificing one regime to optimize another, the sparse architecture maintains distinct expert pathways while sharing common latent structure. The decoded velocity fields retain the tokenizer's mass-conserving constraint, with a physically verifiable velocity divergence of $\sim 2.8 \times 10^{-10}$ (evaluated post-hoc in FP64) on $128^3$ grids.

\subsection{Qualitative Physical Fields}

Finally, we visualize developed-frame physical fields from the source Zarr stores using spatial slices and physically meaningful channels: velocity magnitude for an obstacle-rich V2 continuous-flow sample and velocity magnitude through a porous permeability field. These fields provide qualitative evidence that the mixed corpus contains visually distinct transport regimes: localized high-gradient bottlenecks and wake-like low-speed regions in the continuous-flow sample, and geometrically confined flow paths in the porous-media sample.

\begin{figure}[H]
    \centering
    \includegraphics[width=\columnwidth]{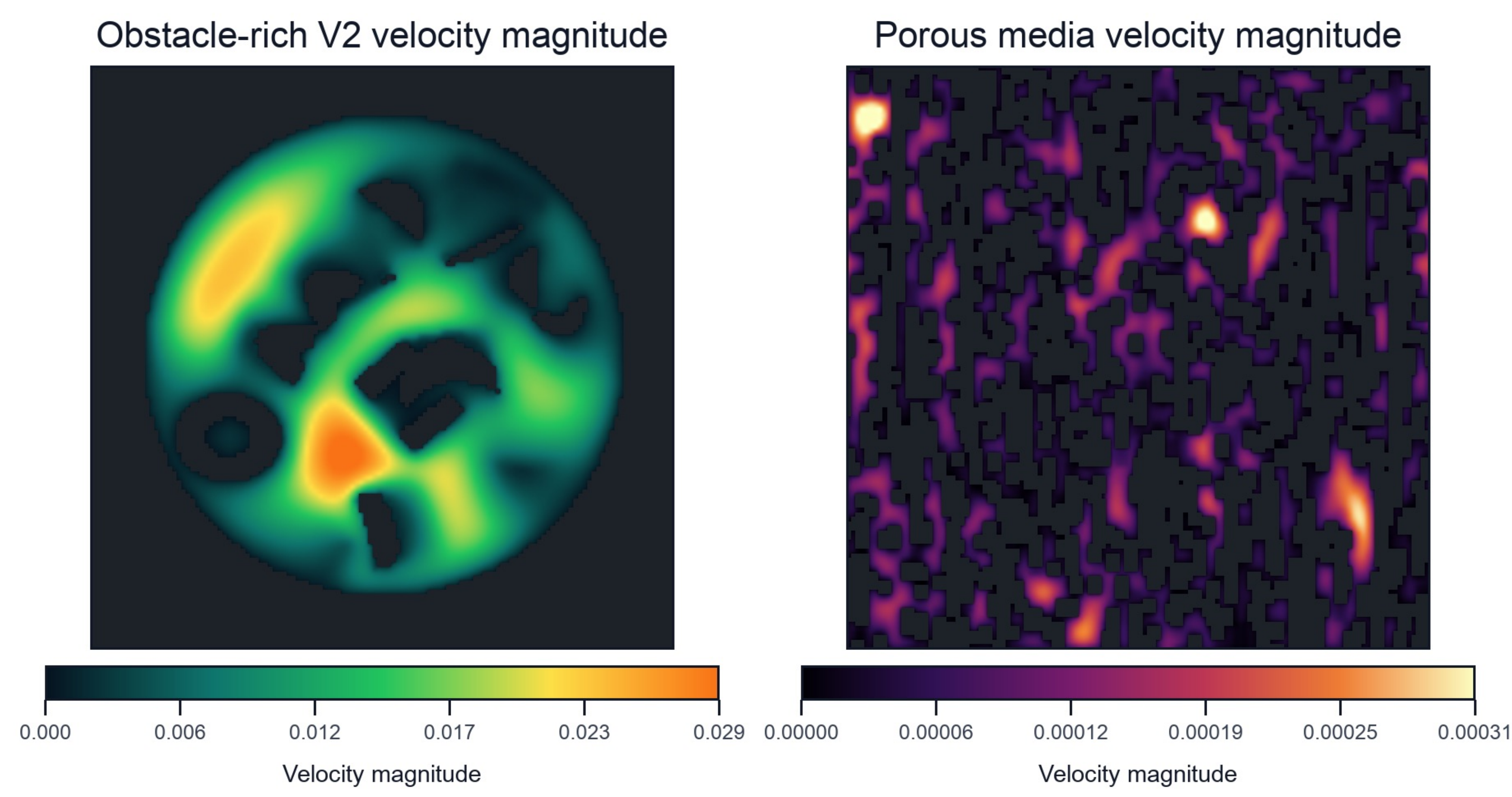}
    \caption{Representative developed physical fields from the mixed corpus. Left: velocity magnitude from an obstacle-rich V2 continuous-flow Zarr store. Right: velocity magnitude from a porous-media permeability sample extracted from the original V3 Zarr store. Spatial slices are shown for both domains.}
    \label{fig:fields}
\end{figure}

\section{Discussion}

The results suggest that sparse conditional computation can provide a practical route around dense multi-physics interference. The router discovers a stable partition in latent space, the experts remain active, and both domains converge to low validation error. This is a meaningful departure from purely compositional approaches: the computation is selected internally by the model rather than assembled externally from a fixed operator library.

At the same time, the experiment should be interpreted carefully. The present evidence establishes routing bifurcation, low validation error, and decoded reconstruction fidelity for representative validation slices.

\section{Conclusion}

Universal physical foundation models must reconcile incompatible differential operators without collapsing into dense compromise. Shodh-MoE addresses this by pairing a physically constrained latent tokenizer with sparse expert routing. The completed 32-H100 run demonstrates autonomous routing bifurcation, active domain-specialized experts, and simultaneous low-error convergence across two contrasting physical regimes. These findings support sparse MoE routing as an architectural bedrock for scalable multi-physics neural operators and generative inverse design.

\section*{Acknowledgments}

This whitepaper reports internal research by Shodh AI. The authors extend their deepest gratitude to the Government of India's IndiaAI Mission. As a supported entity under the IndiaAI Foundational Model Pillar, this research was made possible through their visionary backing and sovereign compute initiatives. We additionally acknowledge the infrastructure, collaboration, and scaling support provided by the NVIDIA Inception program and the NVAITC team, and this research was accelerated via distributed training on NVIDIA H100 architectures.

\end{document}